\documentclass[conference]{IEEEtran}
\usepackage{cite}
\usepackage{amsmath,amssymb,amsfonts}
\usepackage{graphicx}
\usepackage{booktabs}
\usepackage{textcomp}
\usepackage[dvipsnames]{xcolor}

\usepackage{url}
\usepackage{float}

\begin{document}

\title{Beyond Color Geometry: Evaluating Human-Like Color Representations in Vision Models}

\author{\IEEEauthorblockN{Ayan Igali, Pakizar Shamoi}
\IEEEauthorblockA{School of Information Technology and Engineering\\
Kazakh--British Technical University\\
Almaty, Kazakhstan\\
\{ay\_igali, p.shamoi\}@kbtu.kz}}

\maketitle

\begin{abstract}
Do vision models see colors the way humans do? Existing evaluations of color representations usually compare them with geometric spaces such as CIELAB or with discrete color labels. These references capture perceptual distance or category membership, but not the graded way in which people organize colors. We evaluate color grounding against a fuzzy perceptual model with 86 graded categories fitted to human survey data. The framework can be applied to any image encoder and measures three complementary properties: category boundaries, category compactness, and graded alignment beyond what color geometry alone can explain. Across eleven Vision Transformer encoders, the category-level results are broadly similar, whereas graded alignment differs substantially. Masked Autoencoders achieve the strongest beyond-geometry alignment, with confidence intervals that do not overlap those of the other encoders. A layer-wise analysis further shows that masked reconstruction preserves this structure toward the output. On natural images, MAE represents surface color globally, while language-supervised models encode color more strongly in relation to the foreground object. These results show that human-like color grounding has several distinct aspects that should not be reduced to a single score.
\end{abstract}


\section{Introduction}
Modern vision and vision--language models are widely used as feature extractors across computer vision \cite{vit_survey, vlm_survey}. For human-centered and explainable AI, it is useful to ask whether their internal color representations resemble human color perception or mainly reflect pixel-level regularities. Human color perception is not organized as a uniform metric continuum. People group colors into overlapping categories with soft boundaries, and a color can belong partly to several named categories such as red, dark red, gray, or black \cite{continuous_color}.

Most previous studies of color grounding have used geometric color spaces. Abdou et al.~\cite{abdou2021}, for example, showed that text-only language models recover part of the topology of CIELAB \cite{Muratbekova2026}, and later work reported similar results~\cite{patelpavlick2022, sogaard2023}. Such analyses test whether a representation preserves perceptual distances. They do not directly test whether it reproduces the graded category structure of human color perception. Color-naming datasets such as the World Color Survey (WCS)~\cite{wcs} provide human categories, but these are usually treated as hard labels and therefore lose partial membership near category boundaries.

We address this gap with a fuzzy perceptual reference. In a fuzzy color space, each color is represented by graded membership in several human-named categories rather than by a single label. We use COLIBRI~\cite{colibri2025}, which defines 86 categories over Hue, Saturation, and Intensity using membership functions fitted to a large human survey. This lets us compare model representations with human color organization at both categorical and graded levels.

We apply the evaluation to eleven encoders from the Vision Transformer family. 
The main contributions of this paper are:
\begin{itemize}
\item We introduce a reusable framework for evaluating color representations against the fuzzy human color model. The framework applies to any image encoder and reports complementary measures of categorical and graded color organization.

\item We propose a geometry-controlled alignment measure that separates agreement with human color structure from conventional perceptual color distance. 

\item We provide a systematic comparison of eleven Vision Transformer encoders across multiple training regimes, including layer-wise and natural-image analyses


\end{itemize}

The remainder of this paper is structured as follows. Section II reviews related work on color perception, color representations in learned models, and existing approaches to human–model alignment. Section III describes the data, the fuzzy perceptual color model, the evaluated vision encoders, and the experimental methodology. Section IV presents the results. Section V discusses the findings in relation to recent studies. Finally, Section VI concludes the paper.
\section{Related Work}

Research on color categories has long examined the relationship between perception and language, from accounts of basic color terms~\cite{berlinkay} to cross-linguistic analyses based on the World Color Survey~\cite{wcs}. Information-theoretic studies further suggest that color naming reflects efficient communication and that different regions of color space are described with varying precision~\cite{zaslavsky2018}. Across these traditions, human color categories are generally understood as graded and unequal in extent. Colors near a category boundary may partially belong to several categories, while broad categories such as green may occupy substantially more of color space than narrower categories such as yellow. COLIBRI~\cite{colibri2025} models these properties using overlapping fuzzy membership functions defined over Hue, Saturation, and Intensity.

Several studies have investigated whether learned representations recover perceptual color structure. Abdou et al.~\cite{abdou2021} used representational similarity analysis (RSA)~\cite{rsa} and linear probing to show that text-only models encode aspects of CIELAB geometry. Related findings have since been reported for other language models~\cite{patelpavlick2022, sogaard2023}. More broadly, research on human--model alignment has emphasized that alignment should be assessed through multiple complementary measures rather than reduced to a single score~\cite{sucholutsky2023}. Other studies suggest that training data and learning objectives may influence human alignment more strongly than model scale or architectural family~\cite{muttenthaler2023, conwell2024}.

Color has also been evaluated directly in vision and vision-language models. Task-level benchmarks such as ColorBench~\cite{colorbench} assess color perception, reasoning, and robustness through model responses. Such benchmarks measure observable behavior on specific tasks, rather than the organization of color information within the encoder representation. In a different line of work, comparisons between model representations and neural responses indicate that masked autoencoders preserve low-level visual structure while showing weaker correspondence with higher-level neural representations~\cite{karimi2025}.

The studies most closely related to representation-level color evaluation rely either on discrete categories or on continuous perceptual similarity. Akbarinia~\cite{akbarinia2025} evaluates categorical color perception in vision networks using an odd-one-out task over World Color Survey chips, but represents perception through a small set of discrete basic-color labels. Wickramanayaka and Oizumi~\cite{wickramanayaka2025} compare sixteen networks with human color-similarity judgments using RSA and unsupervised Gromov--Wasserstein alignment. Their results suggest that CLIP most closely reproduces the fine-grained geometry of human similarity judgments. Fuzzy color naming has also been studied directly~\cite{benavente2008}, primarily as a method for assigning linguistic labels to colors rather than as a reference for evaluating internal model representations.

Existing approaches therefore evaluate learned color representations against geometric color spaces, continuous similarity judgments, discrete perceptual categories, or task-level behavior. These references capture different aspects of color perception, but none simultaneously represents categorical structure, graded membership, and overlap between linguistic color categories at the representation level (Table~\ref{tab:approaches}).

Our work addresses this gap by evaluating frozen image-encoder representations against COLIBRI, a human-derived fuzzy color model. In contrast to geometric references~\cite{abdou2021} and continuous similarity judgments~\cite{wickramanayaka2025}, COLIBRI describes how colors belong to overlapping linguistic categories with different degrees of membership. At the same time, our partial-correlation analysis controls for geometric color distance, allowing categorical agreement beyond color geometry to be measured explicitly. 

\begin{table}[t]
\caption{Approaches to evaluating color in learned models.}
\label{tab:approaches}
\centering
\scriptsize
\setlength{\tabcolsep}{4pt}
\begin{tabular}{lcccc}
\toprule
Approach & Ground truth & Cat. & Graded & Repr. \\
\midrule
Abdou~\cite{abdou2021} & CIELAB geometry & --- & (cont.) & \checkmark \\

Wickr.~\cite{wickramanayaka2025} & similarity judg. & --- & (cont.) & \checkmark \\

Akbarinia~\cite{akbarinia2025} & hard categories & \checkmark & --- & \checkmark \\

ColorBench~\cite{colorbench} & task answers & \checkmark & --- & --- \\

Benavente~\cite{benavente2008} & fuzzy naming & \checkmark& \checkmark & --- \\

\textbf{Ours} & fuzzy human (COLIBRI) & \checkmark & \checkmark & \checkmark \\
\bottomrule
\end{tabular}
\end{table}

\section{Methods}
\begin{figure*}[t]
\centering
\includegraphics[width=0.90\textwidth]{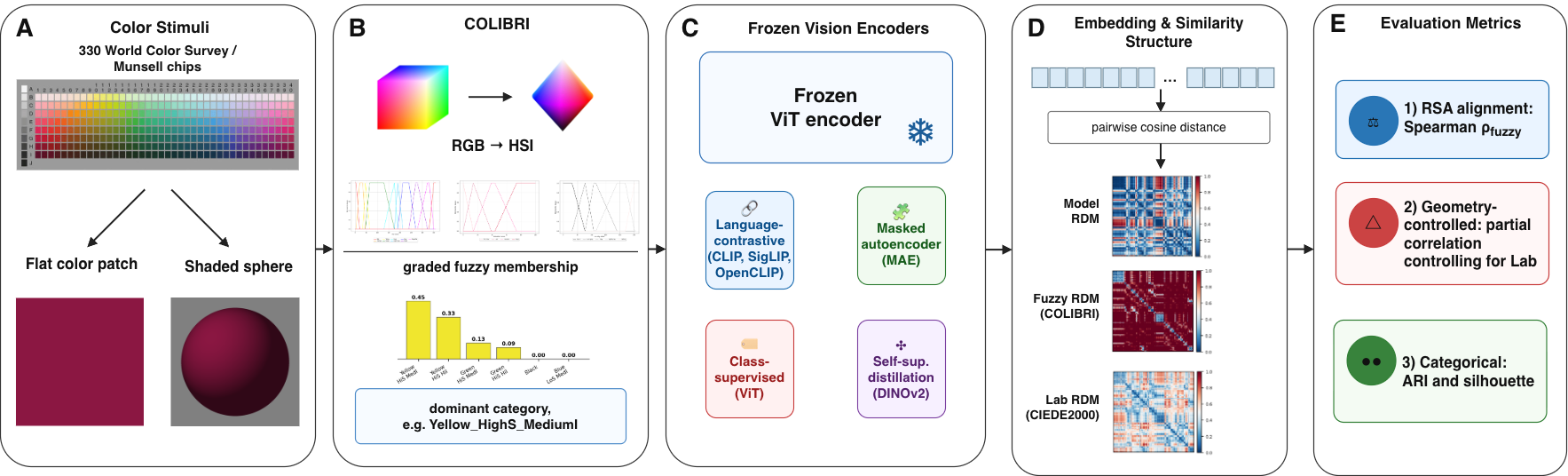}
\caption{
Overview of the evaluation pipeline. 
World Color Survey / Munsell chips are rendered as flat patches and shaded spheres, 
mapped to COLIBRI fuzzy human color memberships, passed through ViT encoders, 
and evaluated by comparing model representational structure with fuzzy, geometric, and categorical color structure.
}
\label{fig:pipeline}
\end{figure*}

We evaluate color grounding by comparing the structure
of a model’s embedding space with the human color struc-
ture represented by COLIBRI. The framework requires only
forward passes through a frozen encoder and reports several complementary measures. Figure \ref{fig:pipeline} summarizes the evaluation
pipeline.

\subsection{Data}
We use the 330 Munsell chips from the World Color Survey \cite{wcs}, a standard stimulus set in color-cognition research, together with their CIELAB coordinates under illuminant C. 

Uniform color fields differ from the natural images on which the encoders were trained, so we render each chip at $224 \times 224$ as a flat patch and a shaded sphere. We use shaded spheres as an intermediate stimulus between uniform color patches and natural images. Compared with flat patches, they add shading, contrast, shape, and spatial variation while keeping the underlying surface color controlled. 
COLIBRI provides the human reference for both forms. Its 86-dimensional membership vectors are used for continuous analyses, and the dominant categories are used for clustering. CIEDE2000 \cite{ciede2000} distances in CIELAB serve only as a geometric control used to test whether model--COLIBRI alignment remains after physical color geometry is accounted for.

\begin{figure}[t]
\centering
\includegraphics[width=0.95\columnwidth]{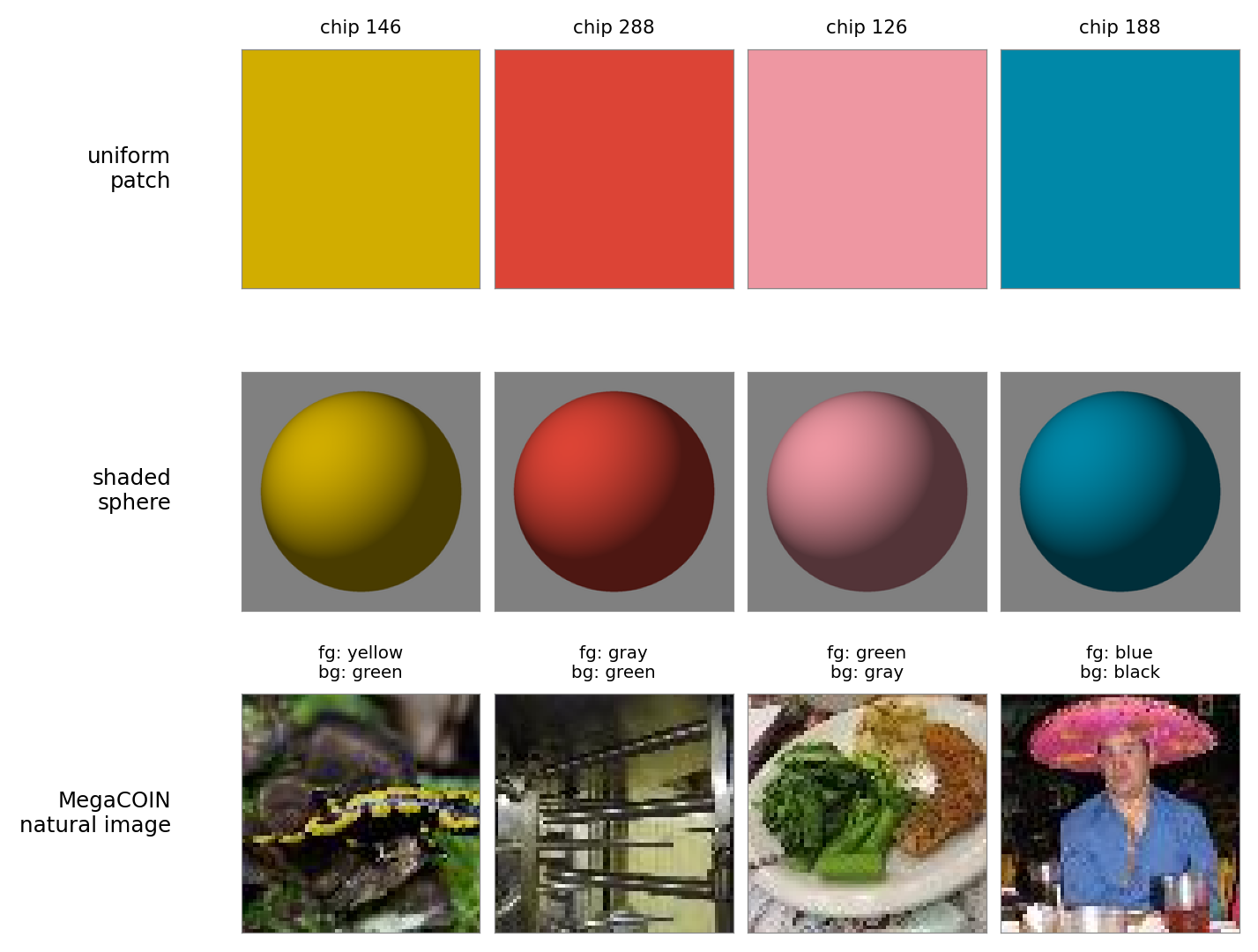}
\caption{Representative stimuli used in the experiments.}
\label{fig:stimuli}
\end{figure}

We also use the Tiny ImageNet subset of MegaCOIN~\cite{megacoin}. Each image is annotated with a human color term for the foreground object and another for the background, which allows us to examine color representations in multi-colored scenes. The foreground probe uses $7{,}997$ images balanced across the eleven color terms; the background probe and the paired object-selectivity gap use the $7{,}744$ of these that also carry a valid background label. Representative examples of all three stimulus formats are shown in Fig.~\ref{fig:stimuli}.

\subsection{COLIBRI: A fuzzy perceptual color space}
COLIBRI~\cite{colibri2025} is a fuzzy color model grounded in human color-naming data. Rather than assigning each color to a single label, it represents graded membership in overlapping linguistic categories. An RGB color is first converted to Hue--Saturation--Intensity (HSI), after which membership is computed using triangular or trapezoidal fuzzy sets derived from a survey of 2,496 participants.

\begin{figure*}
    \centering
    \includegraphics[width=0.89\linewidth]{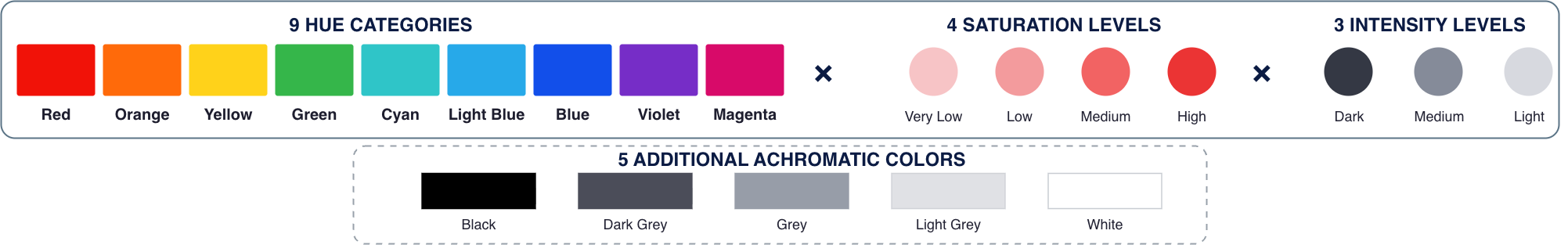}
    \caption{COLIBRI category structure: nine hues, four saturation levels, three intensity levels. At very low saturation a color is named by an achromatic grey level and at extreme intensity by black or white, regardless of hue, giving five achromatic categories. }
    \label{fig:colibri_colors}
\end{figure*}

As shown in Fig.~\ref{fig:colibri_colors}, COLIBRI partitions hue into nine chromatic categories, saturation into four levels, and intensity into three levels. Not all combinations define separate categories, because extreme intensity values map to black or white and very low saturation maps to dark grey, grey, or light grey regardless of hue. The remaining categories combine nine hues, three chromatic saturation levels, and three intensity levels, giving
$9\times3\times3+5=86$ categories.

For a color $(h,s,i)$ and category $v=(v_H,v_S,v_I)$, the composite membership is
$\mu_{v_H}(h)\mu_{v_S}(s)\mu_{v_I}(i)$.
The resulting 86-dimensional vector is used in our continuous analyses, while its largest component defines the dominant label used for clustering. This representation preserves category overlap near perceptual boundaries and reflects human color naming rather than geometric distance alone.
\subsection{Models}
We evaluate eleven encoders spanning four training regimes and two model scales: 
\begin{itemize}
    \item Language-supervised models: CLIP ViT-B/32 and ViT-L/14~\cite{clip}, OpenCLIP ViT-L/14 trained on LAION-2B~\cite{openclip}, SigLIP-Large~\cite{siglip}, and multilingual SigLIP-2-Large~\cite{siglip2}.
    \item Class-supervised models: ViT-B/16 and ViT-L/16 trained on ImageNet-1k~\cite{vit}.
    \item  Reconstruction models: MAE ViT-B/16 and ViT-L/16~\cite{mae}.
    \item Self-distillation models: DINOv2 ViT-B/14 
    and ViT-L/14~\cite{dinov2}.
\end{itemize}

\subsection{Experimental Settings}
All models checkpoints are public and are used without fine-tuning. For image--text models, we use the post-projection image embedding. For MAE, patch masking is disabled and patch tokens are mean-pooled. For supervised ViTs, we use the final class token. Each stimulus produces one embedding, and pairwise dissimilarity is measured with cosine distance.

All models share a ViT backbone but differ in training data, scale, patch size, and recipe, so we report associations with the training regime rather than attributing differences to the objective alone.

\subsubsection{Chip experiments} 
We combine RSA with category-level metrics. For embeddings $\mathbf{x}_i$ of $N$ stimuli, we construct a representational dissimilarity matrix (RDM) using cosine distance:
$D^{\mathrm{emb}}_{ij}=1-\cos(\mathbf{x}_i,\mathbf{x}_j)$.

We compare the upper triangle of the embedding RDM with the upper triangle of a reference RDM using Spearman rank correlation. We denote the correlation with the COLIBRI RDM by $\rho_{\text{fuzzy}}$. The COLIBRI RDM is computed with cosine distance between the 86-dimensional membership vectors.

COLIBRI and CIELAB are related, so a high model--COLIBRI correlation may partly reflect ordinary color geometry. To isolate the remaining graded structure, we compute a partial rank correlation between the model and COLIBRI RDMs while controlling for the CIELAB/CIEDE2000 RDM:
\begin{equation}
p_\Delta
=
\frac{
\rho_{m,f} - \rho_{m,l}\rho_{f,l}
}{
\sqrt{
(1-\rho_{m,l}^{2})(1-\rho_{f,l}^{2})
}
}.
\end{equation}
Here $m$, $f$, and $l$ denote the model, fuzzy (COLIBRI), and Lab RDMs, so that $\rho_{m,f}$ is exactly $\rho_{\text{fuzzy}}$ from above. Applying the standard first-order partial-correlation formula to the Spearman coefficients gives this beyond-geometry measure.

Because the 330 chips act as dominant examples of only 50 of the 86 categories, and some of those categories are represented by very few chips, the clustering metrics (silhouette and ARI) use only categories with at least four chips. The RSA measures instead compare the full pairwise structure of the 86-dimensional membership vectors, so they use all 330 chips regardless of how often each category is dominant.

For category-level analyses, each chip is assigned its dominant COLIBRI label $c(i)$. The silhouette score measures whether embeddings are closer to members of their own category than to members of the nearest alternative category, so higher values indicate more compact categories; it uses the COLIBRI labels directly and involves no clustering. Separately, we run $k$-means on the embeddings ($\texttt{n\_init}=10$) and compare the resulting clusters against the COLIBRI labels with the Adjusted Rand Index (ARI), which is chance-corrected so that random agreement is approximately zero.

Silhouette and ARI describe global category organization. All RDMs use cosine distance except the CIELAB reference, which uses CIEDE2000. We estimate $95\%$ percentile bootstrap confidence intervals by resampling the 330 chips, rather than the individual pairwise distances.


\subsubsection{Natural-image experiment} 
We use the balanced MegaCOIN images, extracting embeddings exactly as for the WCS stimuli and training a separate linear probe for the foreground-object color and for the background color. Each probe uses five-fold cross-validation and is scored by balanced accuracy, which corrects for the imbalance in the background labels. Object selectivity is the gap $\Delta = \mathrm{acc}_{\mathrm{fg}} - \mathrm{acc}_{\mathrm{bg}}$, with confidence intervals from a paired bootstrap over the shared images.

\section{Results}
\begin{figure*}[t]
\centering
\includegraphics[width=0.85\textwidth]{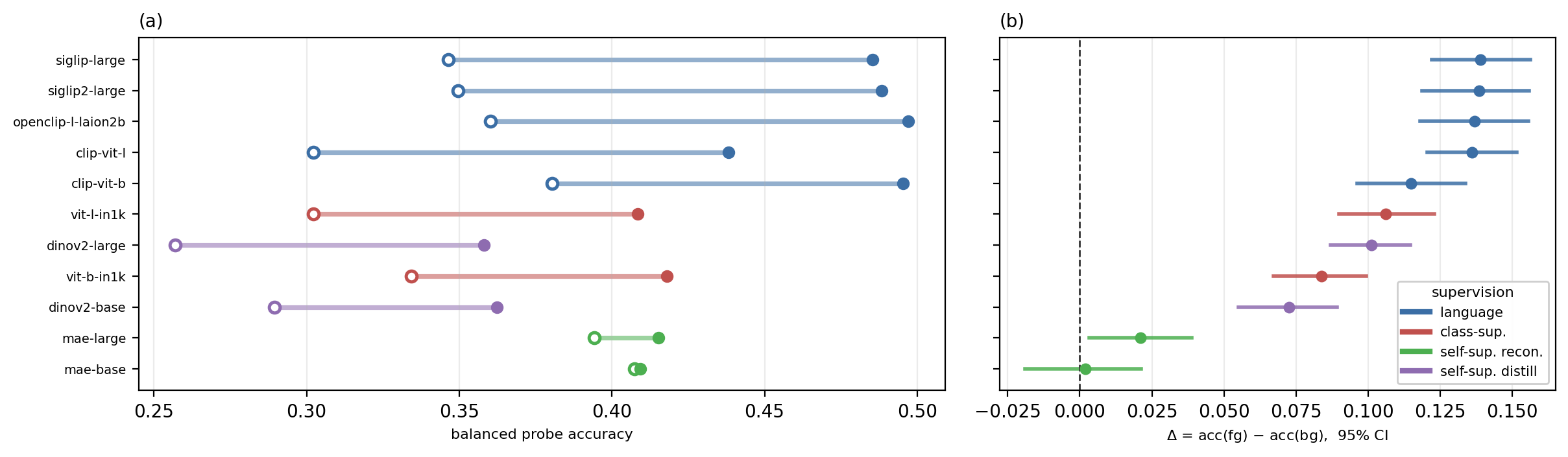}
\caption{
Natural images from Tiny ImageNet.\ (a) Balanced probe accuracy for background ($\circ$) and foreground-object ($\bullet$) color. Segment length shows the object-color selectivity of each encoder.\ (b) The same effect expressed as $\Delta=\mathrm{acc}_{\mathrm{fg}}-\mathrm{acc}_{\mathrm{bg}}$, with paired-bootstrap $95\%$ confidence intervals over $7{,}744$ images.
}
\label{fig:megacoin}
\end{figure*}
\subsection{Color Representation Analysis}
In table \ref{tab:all}, we report results for shaded spheres, which are closer to the encoders' training distribution than uniform patches. 
\begin{table*}[t]
\caption{All eleven encoders, ordered by sphere $\rho_{\text{fuzzy}}$. ARI and silhouette are reported on spheres; the graded measures $\rho_{\text{fuzzy}}$ and $p_\Delta$ are reported on both flat patches (pt) and shaded spheres (sph).
Sphere columns carry $95\%$ bootstrap confidence intervals; patch columns give point estimates.}
\label{tab:all}
\centering
\scriptsize
\setlength{\tabcolsep}{4pt}
\begin{tabular}{llcccccc}
\toprule
& & & & \multicolumn{2}{c}{$\rho_{\text{fuzzy}}$} & \multicolumn{2}{c}{$p_\Delta$} \\
\cmidrule(lr){5-6}\cmidrule(lr){7-8}
Model & Supervision\ & ARI & Sil. & pt & sph & pt & sph \\
\midrule
MAE-large & Reconstruction & 0.358\,[.29,.41] & 0.071\,[$-$.04,.12] & \textbf{0.665} & \textbf{0.673\,[.65,.70]} & \textbf{0.401} & \textbf{0.467\,[.43,.50]} \\
MAE-base & Reconstruction & 0.318\,[.29,.41] & 0.094\,[$-$.01,.13] & \textbf{0.661} & \textbf{0.667\,[.64,.69]} & \textbf{0.384} & \textbf{0.455\,[.42,.49]} \\
DINOv2-base & Self-distillation & 0.348\,[.26,.38] & 0.081\,[.01,.12] & 0.524 & 0.576\,[.55,.60] & 0.298 & 0.248\,[.21,.28] \\
CLIP-B & Language & 0.338\,[.27,.43] & 0.177\,[.10,.22] & 0.400 & 0.563\,[.54,.59] & 0.236 & 0.207\,[.18,.24] \\
ViT-B (IN1k) & Class & 0.285\,[.26,.40] & 0.138\,[.06,.18] & 0.571 & 0.560\,[.54,.59] & 0.289 & 0.237\,[.20,.27] \\
DINOv2-large & Self-distillation & 0.283\,[.25,.36] & 0.067\,[$-$.00,.10] & 0.444 & 0.548\,[.52,.58] & 0.171 & 0.227\,[.19,.26] \\
SigLIP-2-L & Language & 0.331\,[.25,.38] & 0.145\,[.07,.18] & 0.499 & 0.543\,[.51,.57] & 0.245 & 0.277\,[.24,.32] \\
CLIP-L & Language & 0.285\,[.24,.36] & 0.141\,[.07,.18] & 0.394 & 0.533\,[.50,.56] & 0.213 & 0.193\,[.15,.24] \\
SigLIP-L & Language & 0.271\,[.23,.35] & 0.103\,[.03,.15] & 0.506 & 0.476\,[.44,.51] & 0.264 & 0.171\,[.12,.22] \\
ViT-L (IN1k) & Class & 0.328\,[.27,.41] & 0.112\,[.04,.16] & 0.510 & 0.457\,[.43,.49] & 0.169 & 0.163\,[.13,.20] \\
OpenCLIP-L & Language & 0.312\,[.24,.37] & 0.118\,[.05,.16] & 0.352 & 0.444\,[.41,.48] & 0.163 & $-$0.027\,[$-$.08,.02] \\
\bottomrule
\end{tabular}
\end{table*}
The category-level metrics vary little across models. ARI values lie in a narrow band with overlapping bootstrap intervals, and silhouette scores overlap as well, so neither category boundaries nor cluster compactness clearly separates the training regimes.

The graded measures show a different pattern. MAE alignment is high and nearly unchanged across rendering formats, whereas alignment for the language-supervised models rises markedly from flat patches to shaded spheres. Both $\rho_{\text{fuzzy}}$ and $p_\Delta$ place the two MAE models above all other encoders. MAE reaches $\rho_{\text{fuzzy}}=0.67$ $[0.65,0.70]$ and $p_\Delta=0.47$ $[0.43,0.50]$, and on each measure its interval clears the best non-MAE encoder, which reaches only $0.58$ and $0.28$. The same ordering holds on flat patches, where MAE again leads on both $\rho_{\text{fuzzy}}$ (around $0.66$ against $0.57$) and $p_\Delta$ (around $0.39$ against $0.30$), so the advantage survives even after Lab geometry is partialled out and does not rest on the naturalistic rendering. This is not explained by tighter clusters, since MAE has relatively low silhouette scores. It reflects instead a closer match to the graded structure of COLIBRI.


\subsection{Natural images: global and object-bound color representations}
\label{sec:natural}


On the MegaCOIN images, the foreground-object color and background color produce different model rankings (Fig.~\ref{fig:megacoin}). Language-supervised models decode the foreground object's color most accurately, whereas the two MAE models perform best on background color. This difference is summarized by object selectivity, $\Delta$. It is largest for language-supervised models ($+0.12$ to $+0.14$), intermediate for class supervision and self-distillation ($+0.07$ to $+0.11$), and close to zero for masked reconstruction ($+0.02$ and $+0.00$). The confidence interval for MAE-base includes zero, and neither MAE interval overlaps the interval of any other encoder. Embedding dimension does not explain the pattern: DINOv2-large and MAE-large both have 1,024-dimensional embeddings, yet DINOv2-large has the lowest background-color accuracy.

These results suggest that masked reconstruction retains a more global representation of surface color, whereas language supervision makes color more specific to the foreground object. This fits the WCS results, where a single-color stimulus plays to the strength of a global color representation.


\subsection{Layer dynamics}
\label{sec:layers}
\begin{figure}[]
\centering
\includegraphics[width=\columnwidth]{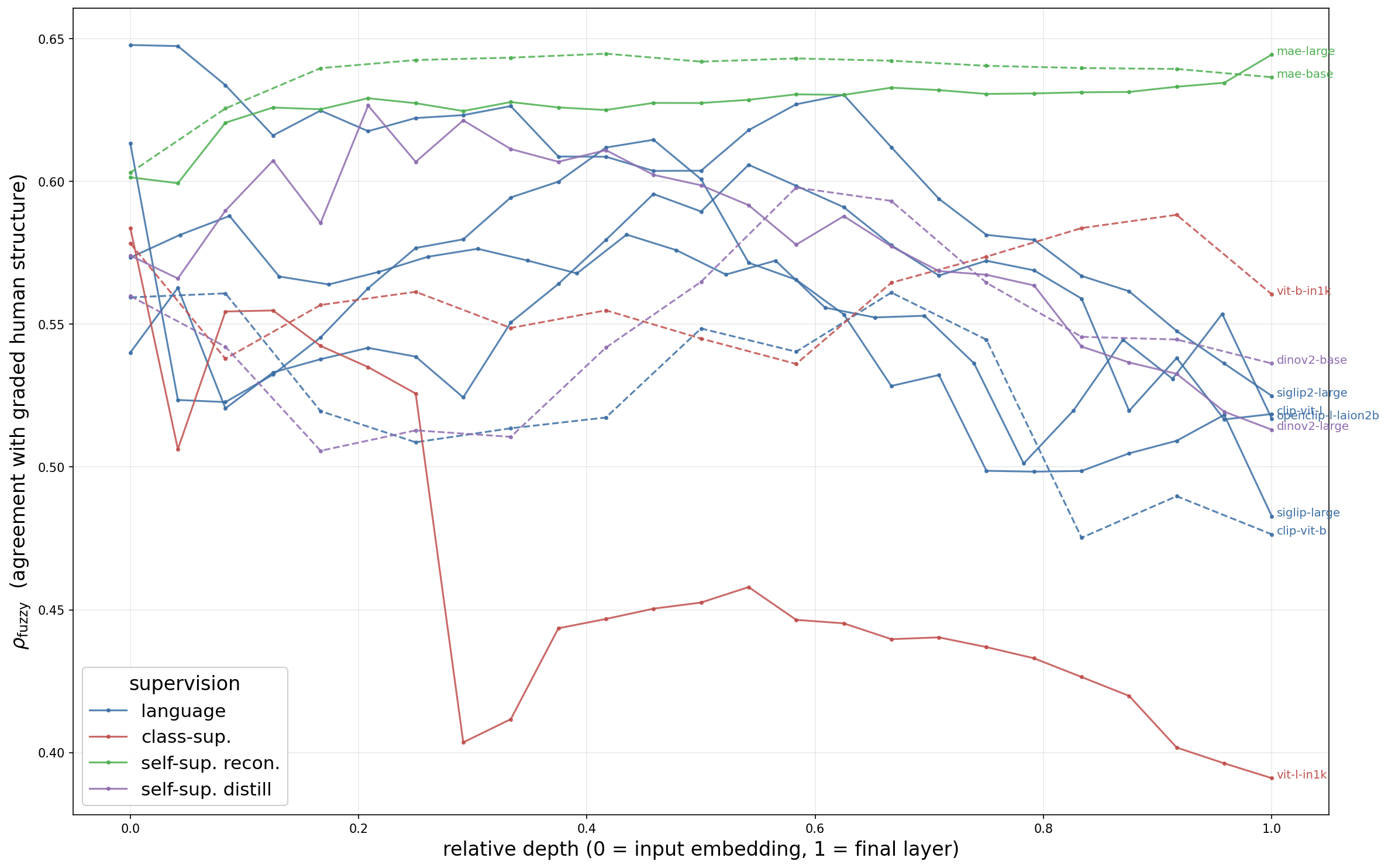}
\caption{Agreement with graded human color structure, $\rho_{\text{fuzzy}}$, across relative ViT depth for all eleven encoders. Color denotes training regime; dashed lines indicate base variants.}
\label{fig:layers}
\end{figure}

The previous analyses use only the final embedding. To trace how graded color structure changes through the network,  we mean-pool the token representations of each transformer layer and recompute $\rho_{\text{fuzzy}}$ for all eleven encoders (Fig.~\ref{fig:layers}). Uniform pooling makes depths comparable within and across models, so these per-layer values are not expected to match the final-embedding scores of Table~\ref{tab:all} exactly; we read the trajectories for their shape, not their absolute level. We report relative depth because the models contain different numbers of layers.

The two MAE models have the highest alignment at nearly every depth and are the only encoders that do not lose alignment toward the output. Language-supervised models begin with high alignment at the input embedding, but it declines steadily through the network. DINOv2 shows a similar downward trend. The class-supervised models differ by scale: ViT-L drops sharply in the middle layers and remains low, whereas ViT-B partially recovers after an initial decline. Thus, graded human color structure is already present in early representations across all models, but masked reconstruction preserves it most consistently through depth.

\section{Discussion}

To better understand the significance of our findings, we compare them with recent studies on human-like color and visual representations in neural networks. 



Across the eleven encoders, category boundaries and cluster compactness look much the same, yet the graded alignment measured after controlling for color geometry pulls them apart, with masked reconstruction standing well above the rest. Since these models also differ in training data, scale, patch size, and recipe, we read this as an association with the training regime rather than a causal effect of the objective on its own. Earlier work points the same way, finding that training data and objectives shape human--model alignment more than architecture does~\cite{muttenthaler2023, conwell2024}.


A plausible reason is that reconstruction has to preserve low-level detail. Color is exactly such a property, so an objective that rebuilds pixels may retain it more faithfully than one organized around semantic classes or image--text matching. This fits evidence that MAEs keep low-level visual structure while diverging from models tuned for higher-level neural representations~\cite{karimi2025}. It stays an interpretation rather than a causal claim, given how many training choices vary at once.

\section{Conclusion}
\label{sec:discussion}
We introduced a framework for evaluating color grounding with a graded fuzzy model of human color categories. The results show why color grounding should not be summarized by a single score. Across eleven ViT encoders, category boundaries and cluster compactness are broadly similar. The models differ more clearly in their agreement with graded human structure after color geometry is controlled. On this measure, the two MAE models achieve the highest scores and are separated from the other encoders by non-overlapping confidence intervals. 
These findings are important because they show that models can organize color categories similarly while still differing in how well they reflect human color perception

The study has several limitations. The 330 WCS chips cover only 50 of the 86 COLIBRI categories as dominant labels, and the stimuli are not statistically independent. The evaluated models differ in data, scale, patch size, training recipe, and objective, so a controlled comparison is needed to isolate the source of the MAE result. The natural-image experiment uses whole-image embeddings and does not include object masks. Finally, all evaluated models belong to the ViT family. Future work should vary the training objective under controlled conditions, include other architectural families, and evaluate object-segmented natural scenes.


\section*{Acknowledgment}
This research has been funded by the Science Committee of the Ministry of Science and Higher Education of the Republic of Kazakhstan (Grant No. AP22786412).

\bibliography{main}

\end{document}